\documentclass[final]{cvpr}

\usepackage{times}
\usepackage{epsfig}
\usepackage{graphicx}
\usepackage{amsmath}
\usepackage{amssymb}
\usepackage{booktabs}
\usepackage[labelfont=bf,small,skip=4pt]{caption}
\usepackage{array,multirow}
\usepackage[table]{xcolor}
\usepackage{inconsolata}
\usepackage{wrapfig}
\DeclareMathOperator{\dec}{Dec}

\def\eg{{\it e.g.}}
\def\ie{{\it i.e.}}
\def\vs{{\it vs.}}

\newcommand{\ignore}[1]{}

\usepackage[pagebackref=true,breaklinks=true,colorlinks,bookmarks=false]{hyperref}

\def\cvprPaperID{521} 
\def\confYear{CVPR 2021}

\begin{document}

\title{Sparse Pose Trajectory Completion}

\author{Bo Liu\\
UC, San Diego\\
{\tt\small boliu@ucsd.edu}
\and
Mandar Dixit\\
Microsoft\\
{\tt\small mandarddixit@gmail.com}
\and
Roland Kwitt\\
University of Salzburg, Austria\\
{\tt\small roland.kwitt@gmail.com}
\and
Gang Hua\\
Wormpex AI Research\\
{\tt\small ganghua@gmail.com}
\and
Nuno Vasconcelos\\
UC, San Diego\\
{\tt\small nuno@ece.ucsd.edu}
}

\maketitle

\begin{abstract}
    We propose a method to learn, even using a dataset where
    objects appear only in sparsely sampled views (e.g. Pix3D), the ability to synthesize a pose trajectory for an arbitrary reference image.
    This is achieved with a cross-modal pose trajectory transfer mechanism. First, a domain transfer function is trained to predict, from an RGB image of the object, its 2D depth map. Then, a set of image views is generated by learning to simulate object rotation in the depth space. Finally, the generated poses are mapped from this latent space into a set of corresponding RGB images using a learned identity preserving transform. This results in a dense pose trajectory of the object in image space. For each object type (\eg, a specific Ikea chair model), a 3D CAD model is used to render a full pose trajectory of 2D depth maps. In the absence of dense pose sampling in image space, these latent space trajectories provide cross-modal guidance for learning. The learned pose trajectories can be {\it transferred\/} to unseen examples, effectively synthesizing all object views in image space. Our method is evaluated on the Pix3D and ShapeNet datasets, 
        in the setting of novel view synthesis under sparse pose supervision, demonstrating 
        substantial improvements over recent art.
\end{abstract}

\vspace{-0.4cm}
\section{Introduction}

It is known that an object viewed from varying angles spans a
manifold in the space of images. Characterization of these manifolds
is important for many problems in computer vision, including
3D scene understanding and view invariant object recognition. However,
these manifolds are quite difficult to learn. This is, in part,
because most object datasets do not contain a dense sampling of object views.

Datasets such as ImageNet or COCO, favor diversity
of instances per object class over diversity with respect to object views.
The situation is different for synthetic datasets, such as
ModelNet~\cite{wu20153d} or ShapeNet~\cite{chang2015shapenet}, which include large numbers
of views per object instance. There is, however, a large domain gap between synthetic datasets
and datasets of natural images. Models trained on synthetic images, therefore, cannot
be deployed on natural images directly.

\begin{figure}[t]
\setlength{\belowcaptionskip}{-0.5cm}
  \centering
  \includegraphics[scale=0.7]{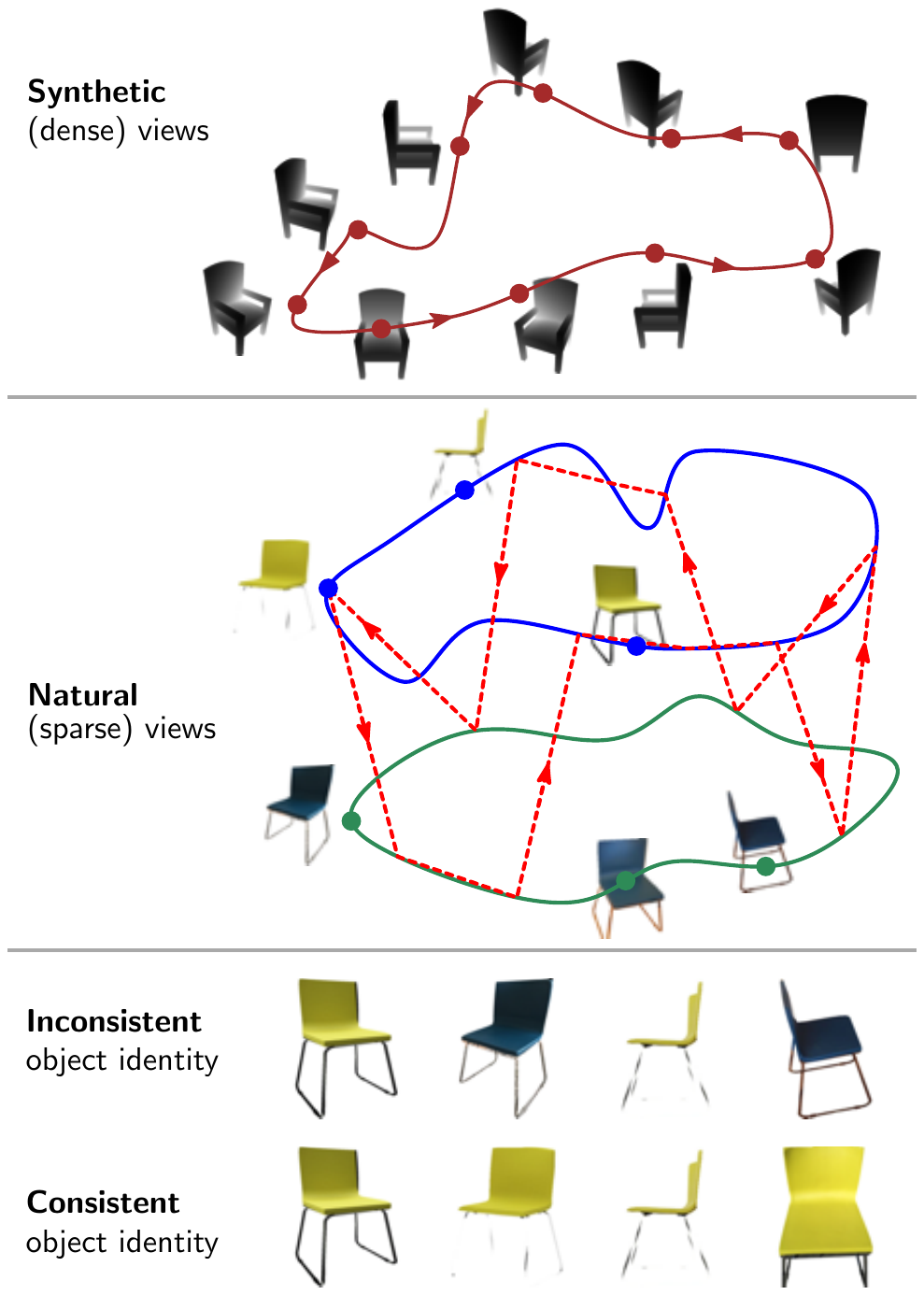}
  \caption{View synthesis via transfer from synthetic data.
    A densely sampled set of synthetic object views is transferred to the
    natural image domain, where only sparse views are available.
    Since synthetic images lack a rich characterization of appearance,
    the transfer can produce a trajectory (shown in \textcolor{red}{red}) that oscillates between
    objects of \emph{similar shape} but \emph{different appearance}. This is likely
    when synthetic views are transferred individually. Consistency of
    object identity requires pose trajectory transfer, \ie, the
    transfer of the entire pose trajectory rather than independent
    views.}
  \label{fig:teaser}
\end{figure}

\begin{figure}[t!]
\setlength{\belowcaptionskip}{-0.5cm}
	\centering
	\includegraphics[width=0.8\linewidth]{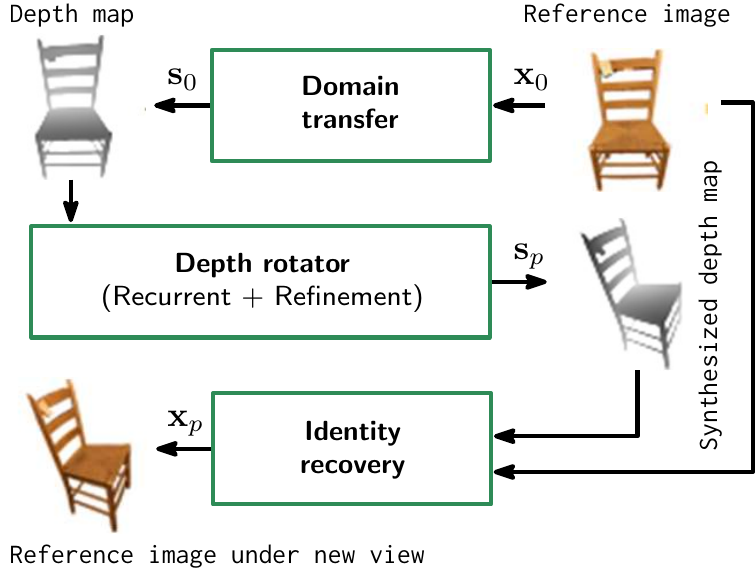}
	\caption{Illustration of the proposed \textbf{DRAW} approach. 
	Arrows show the direction from 
	inputs to outputs.\label{fig:arch}}
\end{figure}

Obtaining natural image datasets that have a large number of instances per object class 
and a dense pose trajectory
per instance, does not seem very feasible at this point. Therefore, a possible solution
to modeling pose manifold could be devised using {\it novel view synthesis\/} 
techniques~\cite{zhou2016view,tatarchenko2016multi} that leverage the synthetic ShapeNet 
dataset and a {\it domain transfer} mapping learned between synthetic and real images~\cite{isola2017image}. 
A view synthesis framework can generate a dense trajectory of object poses, given an input image,
and a domain transfer function could map each generated pose, individually, into the image space. 
Such an approach, however, ignores the problem that the transfer has to preserve object identity across all views. 

This problem is illustrated with an example in Fig.~\ref{fig:teaser} (middle), which shows two objects 
of same shape but different appearance, and their trajectory in image space as a function of view angle. What is also shown is the trajectory spanned by synthetic images from the CAD model of the objects. Since the CAD model does not capture object appearance, the two image domain trajectories map into a single trajectory in the synthetic domain. 
Hence, methods that
rely on synthetic view synthesis and individual view transfer
are likely to oscillate between the synthesis of images of the two objects (\textcolor{black}{red-dashed} trajectory). Image based domain transfer, therefore, does not suffice to solve the natural view synthesis problem. Instead, there is a need for {\it pose trajectory transfer\/} approaches, \ie, methods that transfer entire pose trajectories rather than one view at a time. This problem has some similarities to previous 
work on the hallucination of view changes on scene images~\cite{zhou2016view}. However, such methods assume dense view supervision, which is only available in synthetic or video domains.

In this work, we address the more challenging problem of pose trajectory transfer with {\it sparse natural image data\/}. Unlike ShapeNet, most natural image datasets consist of object instances that are represented by a very few canonical views (\eg, a chair instance captured in $\sim$1-5 poses). Such sparsity of poses, makes it impossible to directly uncover the underlying manifold by training or fine-tuning the existing view synthesis models~\cite{tatarchenko2016multi, yang2015weakly}. 
The pose trajectory must, therefore, be transferred from a densely represented latent space to the sparse image space. 

We propose a formulation where an object's shape space serves as the latent space for view synthesis and transfer. Textureless 3D CAD models are readily available online for many known object classes~\cite{sun2018pix3d} and can be used to sample dense views of the object in the shape space. A trajectory in this auxiliary space can then be used to interpolate between the sparse poses of the object images. More formally, given a reference image $\mathbf{x}_0$, a depth map $\mathbf{s}_0$ is first synthesized. A complete
pose trajectory, ${\bf s}_1, \ldots, {\bf s}_N$, is then generated in the latent space of CAD-based depth maps, and used to provide cross-modal guidance for the modeling of the pose trajectory in image space. 

To guarantee  {\it object identity\/} across the pose trajectory, we introduce the \emph{Domain tRAnsferred vieW synthesis (DRAW)} architecture, as shown in Fig.~\ref{fig:arch}. This consists of three modules.
A {\it domain transfer\/} module is first used to translate the reference image $\mathbf{x}_0$ into the depth map $\mathbf{s}_0$.  A {\it depth rotator\/} is then applied to synthesize a depth map $\mathbf{s}_p$ corresponding to a new view point. Finally, a {\it identity recovery network\/}
is used to generate a new image $\mathbf{x}_p$ of the object under the new viewpoint, based on the rotated depth map and the original image. 
\vskip0.5ex
While the domain transfer and depth rotator modules are variants of the previously studied problems of image translation and synthetic view synthesis, respectively, identity recovery poses a new challenge, critical
to the generation of pose trajectories of consistent object identity. As shown in Fig.~\ref{fig:arch}, it requires the disentanglement  of the shape and appearance components of the reference image $\mathbf{x}_0,$ and the combination of the appearance information with the synthetically
generated new view $\mathbf{s}_p$ of the object shape. 

To achieve this goal, we introduce a new {\it identity recovery network} that takes, as input, the reference image $\mathbf{x}_0$ and the rotated depth map
$\mathbf{s}_p$ and predicts object views under
all various combinations of domain (images \vs~depth) and view
angle (reference \vs~$p^{th}$ view). The requirement for these multiple predictions forces the network to more effectively disentangle shape and appearance information, enabling the synthesis of more realistic views
of the reference object under new poses. This, in turn, enables DRAW 
to synthesize new view points of objects \emph{without} requiring
a training set of images with dense views on pose trajectory. 

\section{Related work}
\label{section:related}

\noindent{\bf Image-to-image transfer.} Transferring images across domains has received substantial attention~\cite{isola2017image,kim2017learning,liu2017unsupervised,zhu2017unpaired}. Novel view synthesis can be considered as a special case of image-to-image transfer, where the source and target represent different views. There are, however, two key differences. First, view synthesis models need to explicitly infer shape from 2D image data. Second, while image transfer usually aims to synthesize style or texture, view transfer needs to ``hallucinate'' unseen shape information.

\vskip0.5ex
\noindent
{\bf Domain adaptation.} Several methods~\cite{panareda2017open,gebru2017fine,zhang2017curriculum,sankaranarayanan2018generate,hoffman2017cycada,tzeng2017adversarial} have been proposed for domain adaptation of visual tasks. Generic domain adaptation aims to bridge the gap between domains by aligning their statistics. DRAW implements a much more complex form of unsupervised domain transfer, leveraging depth information to bridge the natural and synthetic domains and perform bi-directional transfer. Several domain adaptation approaches~\cite{hoffman2016cross,rad2018domain} fuse color and depth features for pose estimation. Unlike these, DRAW relies on image-to-image transfer to leverage viewpoint supervision, which it then uses to decouple appearance and shape. This is critical to recover object identity.

\vskip0.5ex
\noindent
{\bf Novel view synthesis.} Novel view synthesis addresses the generation of images of a given object under new views. One possibility is to simply generate pixels in the target view \cite{tatarchenko2016multi, yang2015weakly}, using  auto-encoders~\cite{tatarchenko2016multi} or recurrent networks~\cite{yang2015weakly}. To eliminate some artifacts of these approaches, Zhou {\it et al.}~\cite{zhou2016view} proposes an appearance flow based transfer module, which reconstructs the target view with pixels from the input and a dense flow map. This, however, cannot hallucinate pixels missing in the source view. Park \etal~\cite{park2017transformation} use an image completion module, after flow based image reconstruction, to compensate for this and \cite{sun2018multi} designs independent modules to predict dense flow and pixel hallucination. 

All these methods require training sets with dense pose trajectories, \ie, large sets of views \emph{of the same object instance}. For example, previous works in~\cite{sun2018multi,tatarchenko2016multi,yang2015weakly,zhou2016view} assume views under 16 or 18 fold azimuth rotation, and the method of Park {\em et al.}~\cite{park2017transformation} requires additional 3D supervision. Hence, existing novel view synthesis methods are usually trained on and applied to ShapeNet~\cite{chang2015shapenet}. Training or fine-tuning these models on the extremely sparse pose trajectories available in natural image datasets does not yield good results. This is shown in our experiments. DRAW makes up for the severe under-representation of image poses with the help of cross-modal supervision. A CAD based object model that generates dense views in the space of depth maps helps to transfer this trajectory to the space of natural images. The only other example of novel view synthesis on natural images is the work of~\cite{geiger2013vision} using the KITTI dataset. However, the generated views are restricted to a few frames and view point supervision is required. In contrast, DRAW generates a large set of views and can be deployed without pose supervision.

\vskip0.5ex
\noindent
{\bf Human pose transfer.} In this setting, the goal is to transfer
a person across poses. Some recent works \cite{ma2017pose,zhao2018multi,neverova2018dense} have addressed this task, leveraging the availability of multi-pose datasets such as DeepFashion~\cite{liu2016deepfashion}. However, besides view points, these methods assume key point supervision~\cite{ma2017pose} or leverage~\cite{neverova2018dense} pre-trained dense human pose estimation networks~\cite{alp2018densepose}. In summary, all these methods require additional supervision and are only  applicable to human pose.

\vskip0.5ex
\noindent
{\bf Single image 3D reconstruction.} Many recent works have proposed to extract 3D shape from a single 2D image. With the availability of large-scale 3D CAD datasets, such as ShapeNet~\cite{chang2015shapenet} and Pix3D~\cite{sun2018pix3d}, remarkable results have been achieved on this task~\cite{wu2017marrnet,choy20163d,fan2017point}. Extraction of 3D shape can be useful for our purpose since an extracted 3D model can be trivially used to generate dense views of the object in the depth space. The depth rotation and refinement module (see section \ref{section:module}) seems to implement this implicitly as it simulates rotation of the object in  depth space. An important distinction, however, between these methods and DRAW, is that the latter does not use any explicit 3D supervision for training.

\section{DRAW}

\subsection{Architecture overview}

The problem can be seen as one of domain adaptation, where
data from a \emph{source domain} (CAD-based depth maps), $\mathbb{S}$, for
which view point annotations are available, is used to improve
the performance of a task (view synthesis) in  a \emph{target domain} 
(images), $\mathbb{T}$, where this is inaccessible. As illustrated in
Fig.~\ref{fig:arch}, this allows the decomposition of the view generation
problem into simpler tasks: a domain adaptation component, which
maps images into depth maps and vice-versa, and a geometric component,
implemented as a 3D rotation of the object. We propose three new
modules to implement these tasks: a \emph{domain transfer} module, a 
\emph{depth rotator} and an \emph{identity recovery} module. 

The domain transfer module, ${\cal F}$, establishes a mapping from 
the target domain $\mathbb{T}$ of RGB images to the
source domain $\mathbb{S}$ of depth maps. It  implements
\begin{equation}
\setlength\abovedisplayskip{6pt}
{\cal F}: \mathbb{T} \to \mathbb{S}, \quad
\mathbf{x}_0 \mapsto {\cal F}(\mathbf{x}_0) = \mathbf{s}_0\enspace,
\setlength\belowdisplayskip{6pt}
\end{equation}
where $\mathbf{x}_0$ and $\mathbf{s}_0$ are a reference image and
a depth map of identical azimuth angle, respectively.
The depth rotator module implements
\begin{equation}
\setlength\abovedisplayskip{1pt}
{\cal G}(\mathbf{s}_0, p) = \mathbf{s}_p
\setlength\belowdisplayskip{6pt}
\end{equation}
for $p = 1, \ldots, N-1$. It  
takes the depth map $\mathbf{s}_0$ associated with the reference view and 
synthesizes the depth maps associated with all other $N-1$ views. This
is realized by two submodules. A recurrent rotator that generates novel
depth map views and a refinement operator that leverages information from
all synthesized depth maps to refine each of them. 
Finally, the identity recovery module implements
\begin{equation} \label{eq:identityrecovery}
\setlength\abovedisplayskip{6pt}
{\cal H}: \mathbb{T} \times \mathbb{S} \to \mathbb{T}, \quad
(\mathbf{x}_0,\mathbf{s}_p) \mapsto {\cal H}(\mathbf{x}_0, \mathbf{s}_p) = \mathbf{x}_p\enspace,
\setlength\belowdisplayskip{6pt}
\end{equation}
taking, as input, the reference view $\mathbf{x}_0$ and the synthesized depth 
map $\mathbf{s}_p$ to produce the synthesized view
$\mathbf{x}_p \in \mathbb{T}$. 
As the name suggests, this modules aims to recover the identity 
of $\mathbf{x}_0$ under the view of $\mathbf{s}_p$. 

\vskip0.5ex
\noindent
\textbf{Domain transfer (DT).}
To learn the domain transfer model $\cal F$, we assume the existence 
of a dataset
with paired images and depth maps, such as Pix3D~\cite{sun2018pix3d} or 
RGB-D~\cite{lai2011large}. This makes the learning of this module a fairly
standard domain transfer problem, where the input is a RGB image
and the output is a depth map. We rely on an image style transfer model, 
similar to~\cite{isola2017image}, to perform the transfer. 
Essentially, it is a fully convolutional
network implemented with ResNet blocks, outputting a depth map and a 
foreground mask that identifies pixels associated with the object.
The object depth map is then obtained by a combination of the two. 
Experimentally, we found that the use of the mask enables cleaner 
depth maps, which eventually lead to better depth rotation results.
We refer the reader to the suppl. material for full architecture
details.

The quality of the synthesized depth map, ${\cal F}(\mathbf{x}_0)$, 
is assessed by the $L_1$ loss, \ie, $\|\mathbf{s}_0-{\cal F}(\mathbf{x}_0)\|_1$.  
Under the framework of~\cite{isola2017image},  this is complemented 
with an adversarial loss that discriminates between synthesized and real 
depth maps. This adversarial loss is implemented with a pair-wise 
discriminator $D$ between the real ($\mathbf{s}_0$) and synthesized 
depth maps, conditioned on $\mathbf{x}_0$. The module
is trained by iterating between learning of the discriminator/critic, with loss
\begin{equation}
\setlength\abovedisplayskip{4pt}
\begin{split}
\label{eq:dr}
{\cal L}^{\texttt{critic}}_{\texttt{DT}}(D) = 
	& ~\mathbb{E}_{\mathbf{x}_0,\mathbf{s}_0}\big[(1-D(\mathbf{x}_0, \mathbf{s}_0))^2\big] ~+ \\  
	& ~\mathbb{E}_{\mathbf{x}_0}\big[(D(\mathbf{x}_0, {\cal F}(\mathbf{x}_0)))^2\big]\enspace.
\end{split}
\setlength\belowdisplayskip{4pt}
\end{equation}
and learning of the mapping ${\cal F}$, with loss
\begin{equation}
\setlength\abovedisplayskip{4pt}
\begin{split}
{\cal L}_{\texttt{DT}}(\mathcal{F}) = & ~\mathbb{E}_{\mathbf{x}_0,\mathbf{s}_0}\big[\|\mathbf{s}_0-{\cal F}(\mathbf{x}_0)\|_1\big]~+ \\
						  &~\lambda_\mathcal{F}~\mathbb{E}_{\mathbf{x}_0}\big[(1-D(\mathbf{x}_0, {\cal F}(\mathbf{x}_0)))^2\big]\enspace,
\end{split}
\setlength\belowdisplayskip{4pt}
\end{equation}
where $\lambda_\mathcal{F}$ is a multiplier balancing the importance of the two
loss components. We have found that the addition of the adversarial
loss helps enforce both sharpness of the output and consistency between 
input and output; consequently, we use this approach for learning all modules
of DRAW.

\vskip0.5ex
\noindent
\textbf{Depth rotation \& refinement (DR).}
The introduction of depth as an intermediate representation for image 
translation transforms view rotation into a geometric operation that can be
learned from datasets of CAD models. 
Rather than reconstructing pixel depths from an appearance map, \eg, using a 
dense appearance flow model~\cite{zhou2016view}, novel depth views are 
synthesized from a reference depth view $\mathbf{s}_0$. 
This leverages the fact that CAD datasets have many views per object and 
the view angles are known. The generation of novel depth views
is implemented with the combination of (1) a depth map generator and (2) 
a 3D refinement module. The \emph{depth map generator}, $\mathcal{G}_1$, is based on a recurrent network,
which takes the reference depth map $\mathbf{s}_0$ as input and
outputs a sequence of depths maps, \ie, 
\begin{equation}
\setlength\abovedisplayskip{6pt}
  \mathbf{s}_p = {\cal G}_1(\mathbf{s}_0, p), \quad
  p = 1, \ldots, N-1\enspace,
\setlength\belowdisplayskip{6pt}
\end{equation}
where $p$ is the azimuth angle. Our implementation of ${\cal G}_1$ is 
based on the ConvLSTM with skip connections of~\cite{sun2018multi}. 
Given a set of depth maps $\{\mathbf{s}_0, \mathbf{s}_1,\ldots,
\mathbf{s}_{N-1}\}$ from $N$ view points, the recurrent generator aims to minimize
\begin{equation}
\setlength\abovedisplayskip{6pt}
{\cal L}_{\texttt{RecGen}}(\mathcal{G}_1) = \sum_{p=0}^{N-1}\mathbb{E}_{\mathbf{s}_0}\big[\|\mathbf{s}_p-{\cal G}_1(\mathbf{s}_0, p)\|_1\big]\enspace.
\label{eqn:recgen}
\setlength\belowdisplayskip{6pt}
\end{equation}

The \emph{refinement module} enforces consistency among neighboring views,
via a 3D convolutional network that leverages the information of 
nearby synthesized views to refine each synthesized view. 
The $N$ depth maps synthesized by the rotator are first stacked into a 
3D volume\footnote{$\oplus$ denotes concatenation along the third dimension.} 
$ \mathbf{s}' = [\mathbf{s}'_0~\oplus~\mathbf{s}'_1~\oplus~\ldots~
  \oplus~\mathbf{s}'_{N-1}]$.
To ensure the refinement of the end views, \eg, $\mathbf{s}_{N-1}$, cyclic 
padding is used on the third dimension. 
The volume $\mathbf{s}'$ is then
processed by 
\begin{equation}
\setlength\abovedisplayskip{4pt}
  \mathbf{s}'' = {\cal G}_2(\mathbf{s}')\enspace.
  \setlength\belowdisplayskip{4pt}
\end{equation}
${\cal G}_2$ is implemented by multiple layers of 3D convolutions 
with skip connections, to produce a 3D volume of concatenated refined
depth maps $\mathbf{s}'' = [\mathbf{s}''_0~\oplus~\mathbf{s}''_1~\oplus~\ldots~
  \oplus~\mathbf{s}''_{N-1}]$.
3D refinement is supervised by a $L_1$ loss, \ie, 
\begin{equation}
\setlength\abovedisplayskip{6pt}
{\cal L}_{\texttt{3D}}(\mathcal{G}_2) = \sum_{p=0}^{N-1}\mathbb{E}_{\mathbf{s}''}
\big[\|\mathbf{s}_p-\mathbf{s}''\|_1\big]\enspace.
\label{eqn:refine}
\setlength\belowdisplayskip{6pt}
\end{equation}
This is complemented by an adversarial loss based on a pair-wise volume 
discriminator $D_V$, between the CAD-based depth map volume ($\mathbf{s}$) 
and the synthesized one ($\mathbf{s}''$), conditioned on $\mathbf{s}'$.
The discriminator/critic loss is
\begin{equation} \label{eq:lossds}
\setlength\abovedisplayskip{6pt}
\begin{split}
{\cal L}^{\texttt{critic}}_{\texttt{V}}(D_V) = 
	& ~\mathbb{E}_{\mathbf{s}',\mathbf{s}}\big[(1-D_V(\mathbf{s}', \mathbf{s}))^2\big]~+ \\
	& ~\mathbb{E}_{\mathbf{s}',\mathbf{s}''}\big[(D_V(\mathbf{s}',\mathbf{s}'')^2\big]\enspace,
\end{split}
\setlength\belowdisplayskip{6pt}
\end{equation}
while $\mathcal{G}_1$ and $\mathcal{G}_2$ are supervised by 
\begin{equation} \label{eq:lossg}
\setlength\abovedisplayskip{6pt}
\begin{split}
{\cal L}_{\texttt{DR}}(\mathcal{G}_1,\mathcal{G}_2) = 
	& ~{\cal L}_{\texttt{RecGen}}(\mathcal{G}_1)~+~\lambda_{\texttt{3D}}~{\cal L}_{\texttt{3D}}(\mathcal{G}_2)~+ \\
	& ~\lambda_{\mathcal{G}}~\mathbb{E}_{\mathbf{s}',\mathbf{s}''}
	\big[(1-D_V(\mathbf{s}', \mathbf{s}''))^2\big]\enspace.
\end{split}
\setlength\belowdisplayskip{6pt}
\end{equation}

\vskip0.5ex
\noindent
\textbf{Identity recovery (IR).}
In standard domain transfer problems, the mapping between source and target 
domains is one-to-one. Each example in the
source domain produces a different image in the target domain.
This is not the case for the transfer between images and depth maps since,
as illustrated in Fig.~\ref{fig:teaser}, objects of the same shape can have
different appearance. Hence, the mapping between images and
depth maps is not bijective. This poses no special problems to
our domain transfer module, which implements
a many-to-one mapping. On the other hand, it implies that it is impossible
to recover the object identity uniquely from its depth map. It follows
that, unlike the domain transfer module, the identity recovery model
cannot be implemented with existing domain transfer networks. In
addition to the depth map $\mathbf{s}_p$, this model must also have
access to the reference image $\mathbf{x}_0$, \ie, implement the
mapping $\cal H$ of~Eq.~ \eqref{eq:identityrecovery}.

In a supervised regression setting, this mapping could be learned from
triplets $(\mathbf{x}_0, \mathbf{s}_p, \mathbf{x}_p)$. However,
we are aware of no datasets that could be used for this.
Most image datasets do not contain depth information or view
point labels. In general, it is difficult to even find multiple views of the
same object with viewpoint annotations. In datasets such as
Pix3D or RGB-D, there are at most a few views per object and these
views are not aligned, \ie, they change from object to object.
Hence, $\cal H$ must be learned from
unpaired data. This, however, is significantly more challenging than 
image-to-image transfer because $\cal H$ has to 1) {\it disentangle\/} the 
appearance and shape information of $\mathbf{x}_0$ and 2) {\it combine\/} the 
appearance information with the shape information of $\mathbf{s}_p$.

To enable this, we propose an encoder-decoder architecture. The
encoder {\it disentangles\/} its input into a pair of shape and appearance 
parameters, via a combination of (1) a structure and (2) an 
appearance predictor. The structure predictor implements the mapping 
$\mathbf{p}={\cal P}(\mathbf{x})$ from the input image $\mathbf{x}$ to shape
parameters $\mathbf{p}$, while the appearance predictor implements 
the mapping $\mathbf{a}={\cal A}(\mathbf{x})$ from the input 
image $\mathbf{x}$ to appearance parameters $\mathbf{a}$.
The decoder then {\it combines\/} these parameters into a 
reconstruction on its output, by taking a vector of concatenated 
appearance and shape parameters and decoding this representation 
into an image.

To force disentanglement, we exploit the fact that, while the object
shape is captured by both its image and depth map,  appearance is only
captured by the image. It follows that (1) combining the shape information
derived from domain A with the appearance information derived from domain B
and (2) reconstructing should produce an image of the object in domain B
under the view used in domain A. Hence, using both the image and shape domains
as A and B, it should be possible to synthesize images with the four
possible combinations of domain (image \vs~depth map) and view
(reference \vs~target). By matching each of these four classes of
synthesized images to true images of the four classes, we encourage the network to learn to disentangle and combine the shape and
appearance representations.
\begin{figure}[t!]
\setlength{\belowcaptionskip}{-0.5cm}
	\centering
	\includegraphics[width=0.46\textwidth]{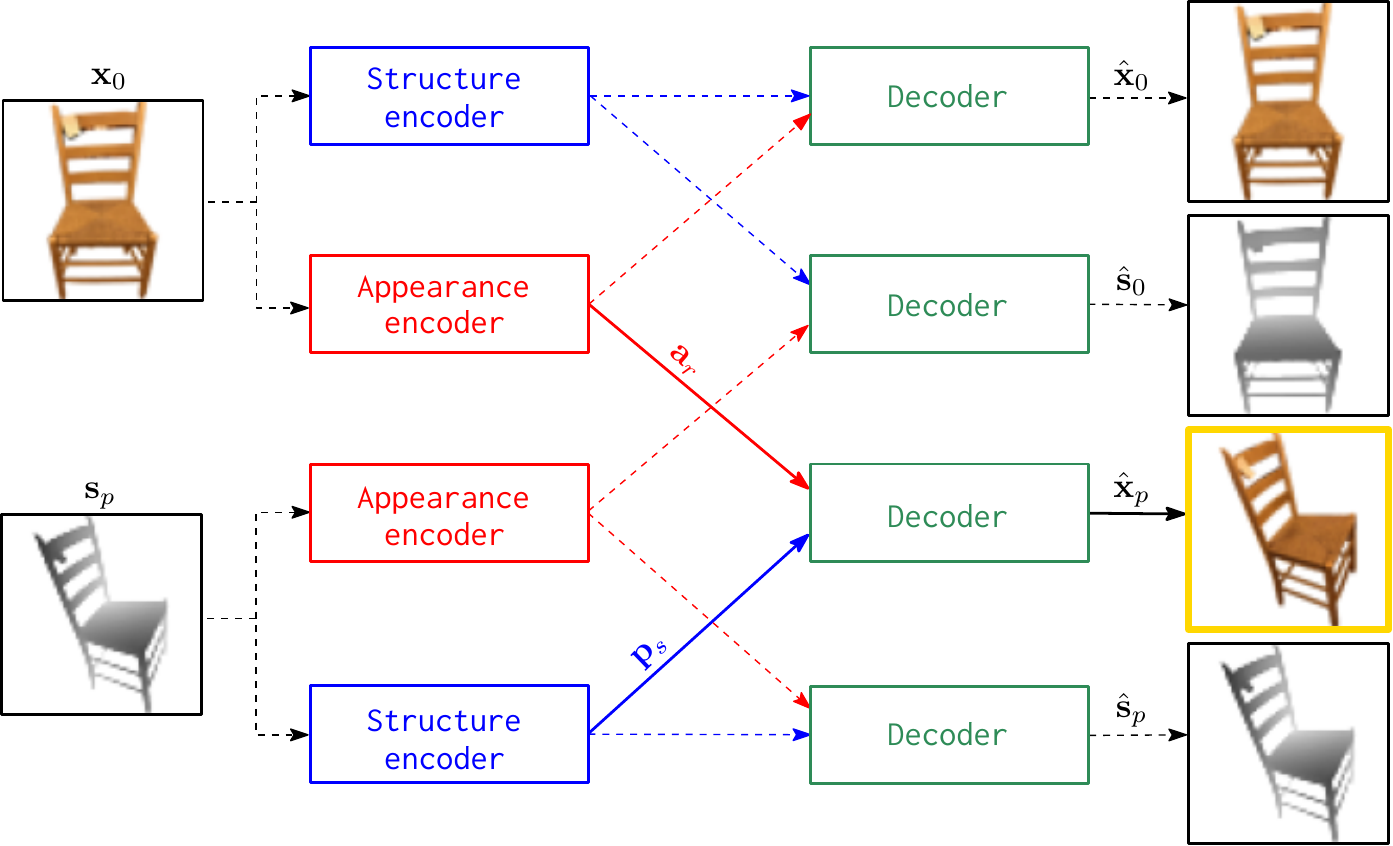}
	\caption{Data flow for the \emph{identity
	recovery} module. Dashed-lines identify the training data flow, solid lines identify the data flow during inference. Blocks of equal color share parameters.}
	\label{fig:identity}
\end{figure}
\begin{figure}[t!]
\setlength{\belowcaptionskip}{-0.25cm}
	\centering
	\includegraphics[width=0.43\textwidth]{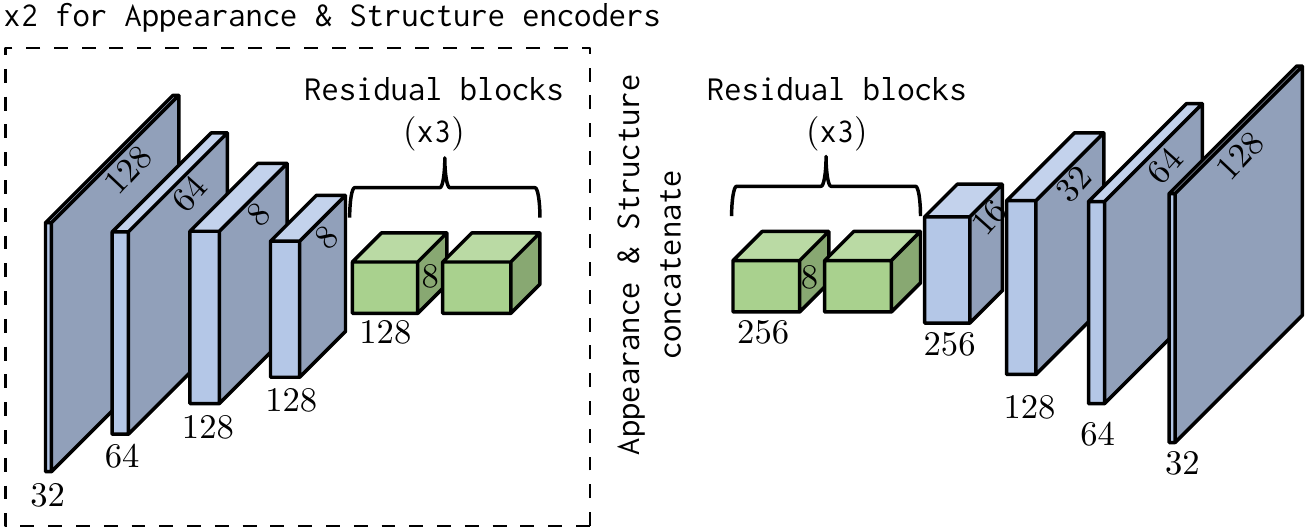}
	\caption{Architecture of the \emph{identity recovery} module.}
	\label{fig:id_structure}
	\vspace{-0.5mm}
\end{figure}

In the multi-view setting, the four combinations are not available, since
$\mathbf{x}_p$ is the target. Yet, the idea can be
implemented with the remaining three combinations:
reference image ($\mathbf{x}_0$), reference depth ($\mathbf{s}_0$)
and target view depth ($\mathbf{s}_p$).
This leads to the architecture of Fig.~\ref{fig:identity},
which combines a pair of encoders and four decoders.
The encoders are
applied to the reference image $\mathbf{x}_0$ and the depth map $\mathbf{s}_p$. This results in two pairs of shape and appearance
parameters
\begin{equation}
\setlength\abovedisplayskip{6pt}
  \mathbf{p}_r = {\cal P}(\mathbf{x}_0), \quad
  \mathbf{a}_r={\cal A}(\mathbf{x}_0)
  \setlength\belowdisplayskip{6pt}
\end{equation}
\begin{equation}
\setlength\abovedisplayskip{6pt}
  \mathbf{p}_s = {\cal P}(\mathbf{s}_p), \quad
  \mathbf{a}_s={\cal A}(\mathbf{s}_p).
  \setlength\belowdisplayskip{6pt}
\end{equation}
The decoders are then applied to the four possible {\it combinations\/} 
of these parameter vectors, synthesizing four images,
\begin{equation}
\setlength\abovedisplayskip{6pt}
  \mathbf{\hat{x}}_0 = \dec(\mathbf{p}_r, \mathbf{a}_r), \quad
  \mathbf{\hat{s}}_0=\dec(\mathbf{p}_r, \mathbf{a}_s),
  \setlength\belowdisplayskip{6pt}
\end{equation}
\begin{equation}
\setlength\abovedisplayskip{6pt}
  \mathbf{\hat{x}}_p = \dec(\mathbf{p}_s, \mathbf{a}_r), \quad
  \mathbf{\hat{s}}_p=\dec(\mathbf{p}_s, \mathbf{a}_s).
\setlength\belowdisplayskip{6pt}
\end{equation}
As shown Fig.~\ref{fig:identity} (\emph{right}), these are all possible combinations
of shape and appearance from the real image with shape and appearance from
the depth map ``image.''  To force the disentanglement into shape and appearance, 
the structure predictors, appearance predictors, and decoders 
share parameters. Note that this implies that only one encoder and one decoder are effectively learned.
During inference, the target image $\mathbf{x}_p$ is obtained with
\begin{equation}
\setlength\abovedisplayskip{6pt}
  \mathbf{\hat{x}}_p = \dec({\cal P}(\mathbf{s}_p),
  {\cal A}(\mathbf{x}_0))\enspace.
\setlength\belowdisplayskip{6pt}
\end{equation}

\begin{figure}[t!]
\setlength{\belowcaptionskip}{-0.5cm}
	\centering
	\includegraphics[width=0.8\columnwidth]{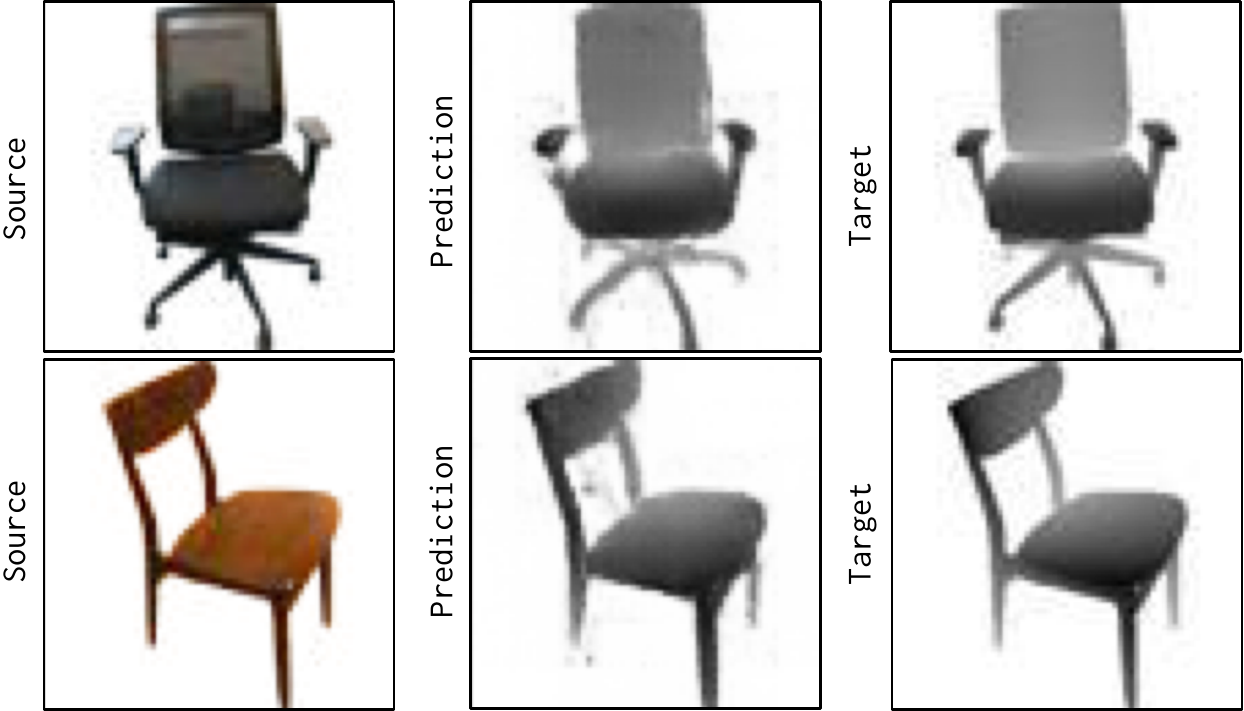}
	\caption{Example results for the \emph{domain transfer} module.}
	\label{fig:domain_results}
\end{figure}
\begin{figure*}[t!]
\setlength{\belowcaptionskip}{-0.25cm}
\centering{
	\includegraphics[width=0.97\textwidth]{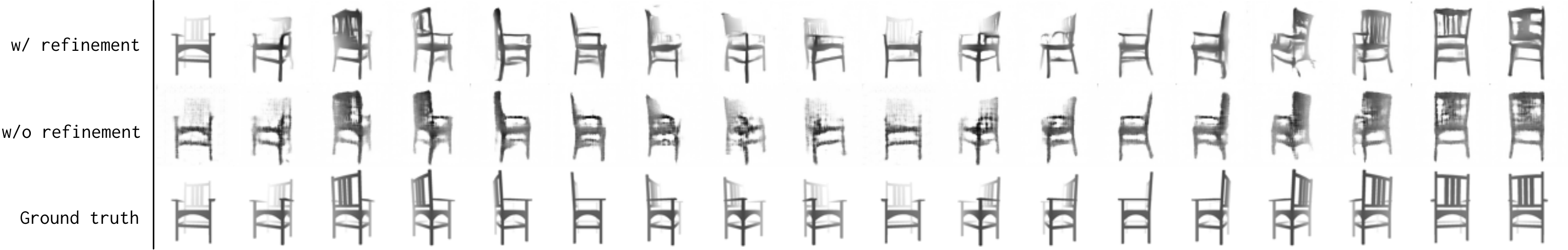}}
	\caption{Qualitative depth rotation results w/ and w/o using 3D refinement, compared to the ground truth.}
	\label{fig:rotate_results}
\end{figure*}
\begin{table*}[t!]
\setlength{\belowcaptionskip}{-0.5cm}
	\begin{footnotesize}
		\centering
		\begin{tabular}{llcccccccccc}
			\toprule
			\textbf{View distance} & & 1 & 2 & 3 & 4 & 5 & 6 & 7 & 8 & 9 & All \\
			\hline
			\multirow{2}{*}{\texttt{w/o refinement}}
			& $L_1$ 	& 0.045  	& 0.049 & 0.049 & 0.050 & 0.049 & 0.049 & 0.048 & 0.047 & 0.047 &  0.048  \\
			& SSIM 		& 0.836 	& 0.813 & 0.808 & 0.803 & 0.801 & 0.803 & 0.806 & 0.825 & 0.839 &  0.813 \\
			\multirow{2}{*}{\texttt{w/ refinement}}
			& $L_1$ 	& \cellcolor{green!20}{0.031} & \cellcolor{green!20}{0.036} & \cellcolor{green!20}{0.037} & \cellcolor{green!20}{0.038} 
						& \cellcolor{green!20}{0.038} & \cellcolor{green!20}{0.037} & \cellcolor{green!20}{0.035} & \cellcolor{green!20}{0.031} 
						& \cellcolor{green!20}{0.026} & \cellcolor{green!20}{0.035} \\
			& SSIM 		& \cellcolor{green!20}{0.945} & \cellcolor{green!20}{0.913} & \cellcolor{green!20}{0.907} & \cellcolor{green!20}{0.898} 
						& \cellcolor{green!20}{0.897} & \cellcolor{green!20}{0.902} & \cellcolor{green!20}{0.915} & \cellcolor{green!20}{0.940} 
						& \cellcolor{green!20}{0.975} & \cellcolor{green!20}{0.918} \\
			\bottomrule
		\end{tabular}
		\caption{Quantitative evaluation of the \emph{depth rotator} module. For $L_1$, lower values are better; for SSIM,  
		higher values are better.}
		\label{tab:rotate_results}
	\end{footnotesize}
\end{table*}

Training of the identity recovery model uses a mix of supervised and unsupervised learning.
Since $\mathbf{x}_0, \mathbf{s}_0,$ and $\mathbf{s}_p$ are available, they provide direct
supervision for the synthesis of the combinations $\mathbf{\hat{x}}_0$, $\mathbf{\hat{s}}_0$, and
$\mathbf{\hat{s}}_p$, respectively. Realized as a a supervised loss, we get
\begin{equation}
\setlength\abovedisplayskip{6pt}
{\cal L}^{\texttt{S}}_{\texttt{IR}} = \mathbb{E}_{\mathbf{x}_0,\mathbf{s}_0,\mathbf{s}_p}[\|\mathbf{x}_0-\mathbf{\hat{x}}_0\|_1 + \|\mathbf{s}_0-\mathbf{\hat{s}}_0\|_1 + \|\mathbf{s}_p-\mathbf{\hat{s}}_p\|_1],
\label{eqn:irdisc}
\setlength\belowdisplayskip{6pt}
\end{equation}
ensuring the quality of {\it both\/} the disentanglement and the image 
synthesis.
This is complemented by an adversarial loss where the combinations
$(\mathbf{x}_0, \mathbf{\hat{s}}_0)$, $(\mathbf{\hat{x}}_0, \mathbf{s}_0)$
and $(\mathbf{\hat{x}}_p, \mathbf{s}_p)$ are all considered as
fake pairs, to be indistinguishable from the real pair
$(\mathbf{x}_0, \mathbf{s}_0)$. This pairwise discriminator/critic
is trained with loss
\begin{equation}
\setlength\abovedisplayskip{8pt}
\begin{split}
{\cal L}^{\texttt{critic}}_{\texttt{IR}}(D) = & ~\mathbb{E}_{\mathbf{x}_0,\mathbf{s}_0}\big[(1-D(\mathbf{x}_0, \mathbf{s}_0))^2\big]~+ \\
				  & ~\mathbb{E}_{\mathbf{x}_0,\mathbf{s}_p}\big[(D(\mathbf{x}_0, \mathbf{\hat{s}}_0))^2\big]~+ \\
   				  & ~\mathbb{E}_{\mathbf{x}_0,\mathbf{s}_0}\big[(D(\mathbf{\hat{x}}_0, \mathbf{s}_0))^2\big]~+ \\
				  & ~\mathbb{E}_{\mathbf{x}_0,\mathbf{s}_p}\big[(D(\mathbf{\hat{x}}_p, \mathbf{s}_p))^2\big]\enspace,
\end{split}
\end{equation}
while the encoder and decoder are trained with loss
\begin{equation}
\setlength\abovedisplayskip{8pt}
\begin{split}
{\cal L}_{\texttt{IR}}(\mathcal{H}) = &~\mathbb{E}_{\mathbf{x}_0,\mathbf{s}_p}[(1-\mathcal{D}(\mathbf{x}_0, \mathbf{\hat{s}}_0))^2]~+ \\
				 					  &~\mathbb{E}_{\mathbf{x}_0,\mathbf{s}_p}[(1-\mathcal{D}(\mathbf{\hat{x}}_0, \mathbf{s}_0))^2]~+ \\
				 					  &~\mathbb{E}_{\mathbf{x}_0,\mathbf{s}_p}[(1-\mathcal{D}(\mathbf{\hat{x}}_p, \mathbf{s}_p))^2]+ 
				 					  {\cal L}^{\texttt{S}}_{\texttt{IR}}  \enspace.
\end{split}
\end{equation}
Fig.~\ref{fig:id_structure} shows the structure of the identity recovery
module.

\vskip0.5ex
\noindent
\textbf{Optimization schedule.}
DRAW is trained in two stages to decouple domain transfer and view point synthesis. 
First, the depth rotation and refinement modules ($\mathcal{G}_1,\mathcal{G}_2)$,  
as well as its discriminator, $D_V$, are optimized with the losses of Eqs.~\eqref{eq:lossds}
and~\eqref{eq:lossg}. 
Once trained, these modules are frozen. The second stage
then addresses end-to-end training of the domain transfer and identity recovery 
modules. The loss
\begin{equation}
\setlength\abovedisplayskip{6pt}
{\cal L}(\mathcal{D}) =  {\cal L}_{\texttt{DT}}^{\texttt{critic}}(D) + \lambda_2{\cal L}_{\texttt{IR}}^{\texttt{critic}}(D)
\setlength\belowdisplayskip{6pt}
\end{equation}
is used to train discriminators/critics, and the loss
\begin{equation}
\setlength\abovedisplayskip{6pt}
{\cal L}(\mathcal{F},\mathcal{H}) = {\cal L}_{\texttt{DT}}(\mathcal{F}) + \lambda_1{\cal L}_{\texttt{IR}}(\mathcal{H})\enspace
\setlength\belowdisplayskip{6pt}
\end{equation}
supervises both parts.

\vspace{-0.5ex}
\section{Experiments}
\label{section:experiments}
We evaluate DRAW using natural images from the Pix3D~\cite{sun2018pix3d} dataset and synthetic images
from the ShapeNet~\cite{chang2015shapenet} dataset.
To ensure enough diversity of instances
in both datasets, we choose two categories: \emph{chairs} and \emph{tables}. First, we evaluate the three modules of DRAW separately on the \emph{chair} class for:
domain transfer between RGB images and depth maps, view synthesis or rotation simulation in the depth space, and 
identity recovery from depth to RGB. The $L_1$ difference norm and structural similarity measure
(SSIM) are used as quantitative synthesis metrics. 
We then compare the performance of the full DRAW framework with three recent view synthesis 
methods~\cite{zhou2016view, tatarchenko2016multi, sun2018multi} on {\it sparse pose completion\/} using the Pix3D dataset and
a subsampled version of the ShapeNet dataset.

\vskip0.5ex
\noindent
\textbf{Datasets.}
On ShapeNet,
$72$ images of size $256\times 256$ pixels are synthesized per CAD model,
using $18$ azimuth angles and elevations 
in $\{0^\circ, 10^\circ, 20^\circ, 30^\circ\}$. 
The dataset is split into $558$ objects for training and
$140$ objects for testing~\cite{zhou2016view}
Pix3D combines 2D natural images with sparse views and 3D CAD models. Images and depth maps are cropped + resized to  $256\times 256$ pixel. 
The dataset includes multiple images aligned with each object. While DRAW does not require this,
they are useful to evaluate identity recovery. 
Training and test sets are split based on objects
to ensure that images with the same object do not appear in training \emph{and} testing. This gives $758$ training images from $150$ 
objects and $140$ test images from $26$ objects.

\subsection{Individual module evaluation} \label{section:module}
\noindent
{\bf Domain transfer (DT).} Since this is a fairly standard module, we do not
carry out a detailed performance evaluation.
Fig.~\ref{fig:domain_results} shows typical domain transfer results. In general, the predicted depth maps are fairly close to the ground-truth.

\vskip0.5ex
\noindent
{\bf Depth rotator (DR).} We compare the proposed combination of rotator +
3D refinement to a variant without the latter. Both models are trained on
the depth maps from $18$ ShapeNet views. Given a reference depth map,
the task is to synthesize the remaining $17$ depth maps. 
Fig.~\ref{fig:rotate_results} shows some typical
examples. Most depths maps are close to the ground truth, but refinement
improves the rendering of fine details. 
Table~\ref{tab:rotate_results} compares the $L_1$ and SSIM scores
of the two methods, with refinement consistently improving the results for both
scores.

\begin{figure}[t!]
\setlength{\belowcaptionskip}{-0.25cm}
	\centering
	\includegraphics[scale=0.6]{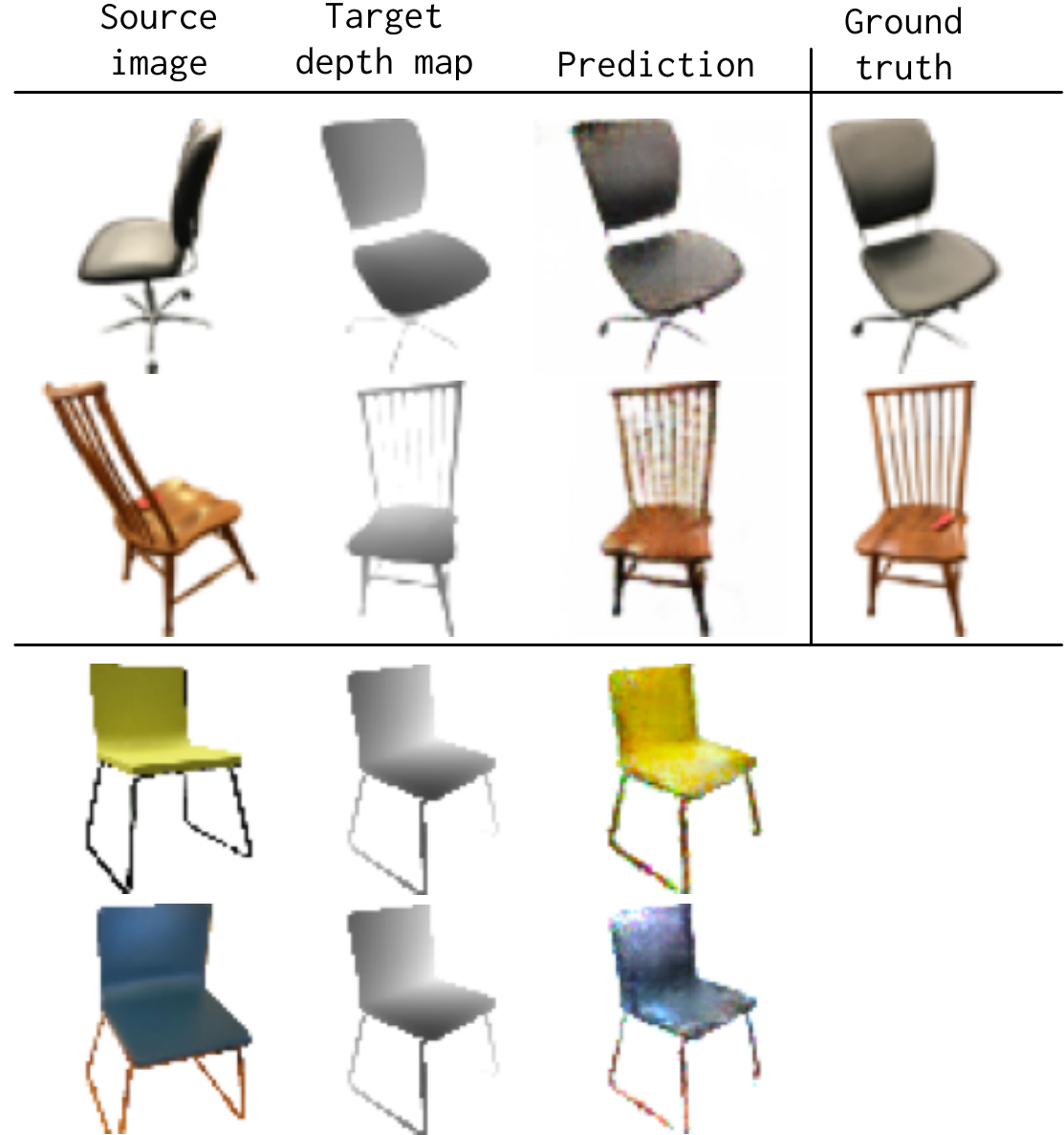}
	\caption{\emph{Identity recovery}, using (1) different target depth maps 
	(\emph{top}) and (2) the same target depth map (due to the sparsity of pose trajectories, we don't have ground truth for the \emph{bottom} part).
	\label{fig:recovery_results}}
\end{figure}

\begin{figure}[t!]
\setlength{\belowcaptionskip}{-0.5cm}
\centering{
\includegraphics[width=1\columnwidth]{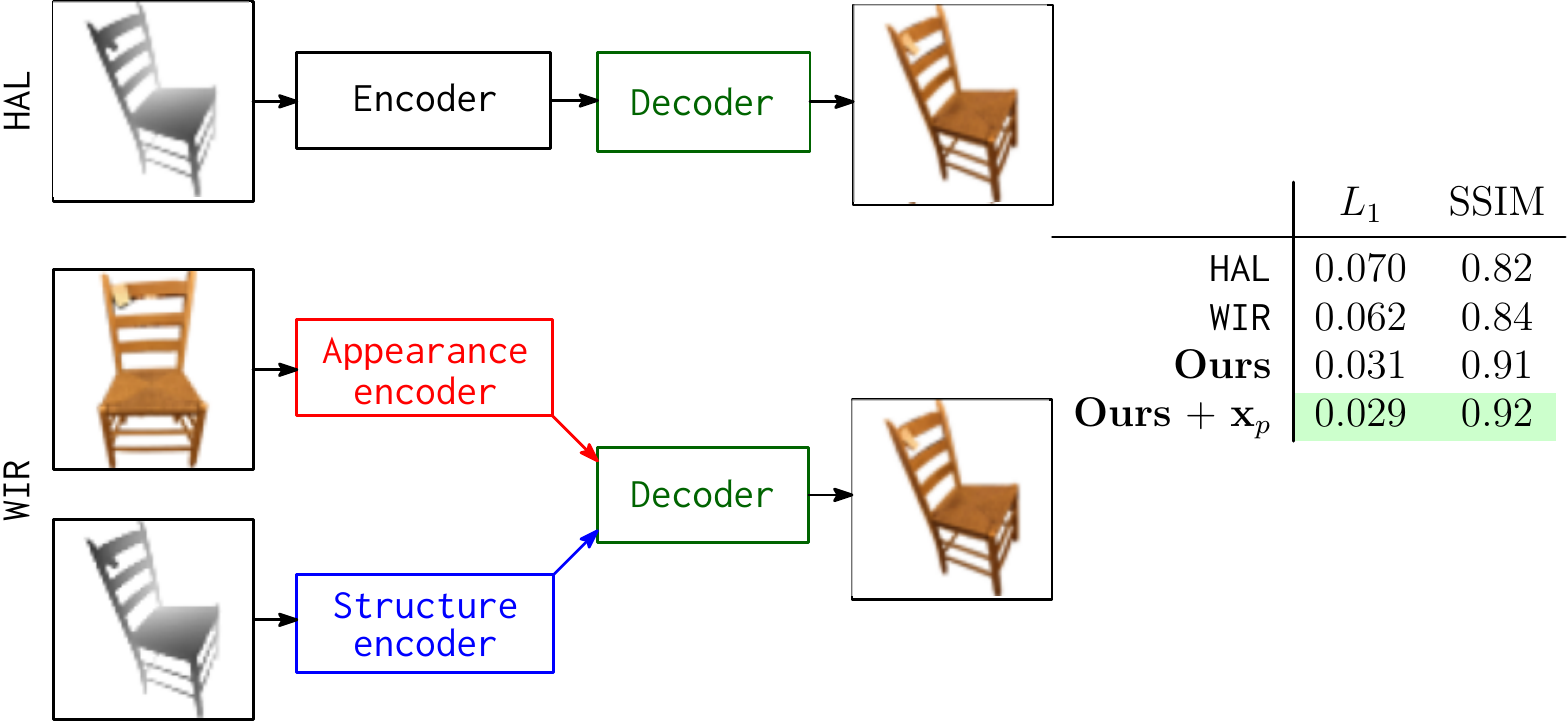}}
	\caption{Comparison of the identity recovery module of Fig.~\ref{fig:identity} to
	(1) a simple image-to-image translation model (\texttt{HAL}, \emph{top}) and (2) a \emph{weak identity recovery} module (\texttt{WIR}, \emph{bottom}), 
	using fewer disentanglement constraints.
	\label{fig:recovery_variation}}
\end{figure}

\begin{figure*}[t!]
\setlength{\belowcaptionskip}{-0.25cm}
\centering{
		\begin{tabular}{c|l}
			\cite{tatarchenko2016multi} & \includegraphics[width=0.85\textwidth]{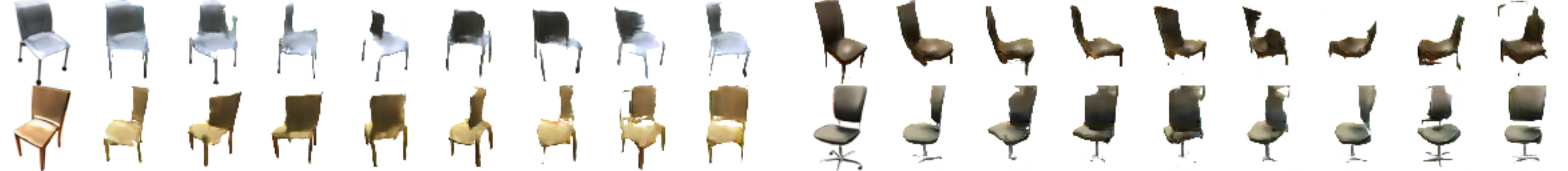} \\
			\hline
			\cite{zhou2016view} & \includegraphics[width=0.85\textwidth]{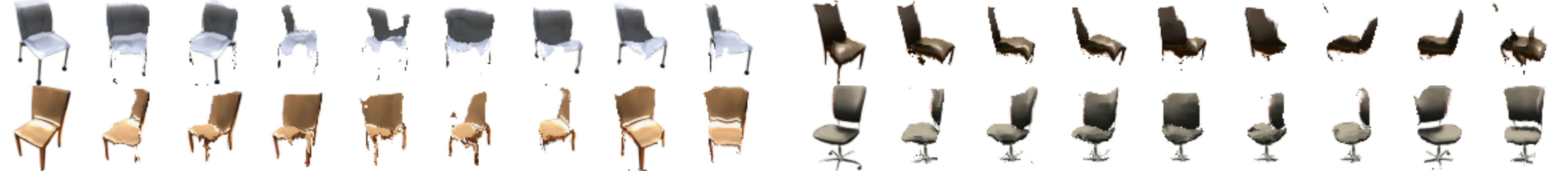} \\
			\hline
			\cite{sun2018multi} & \includegraphics[width=0.85\textwidth]{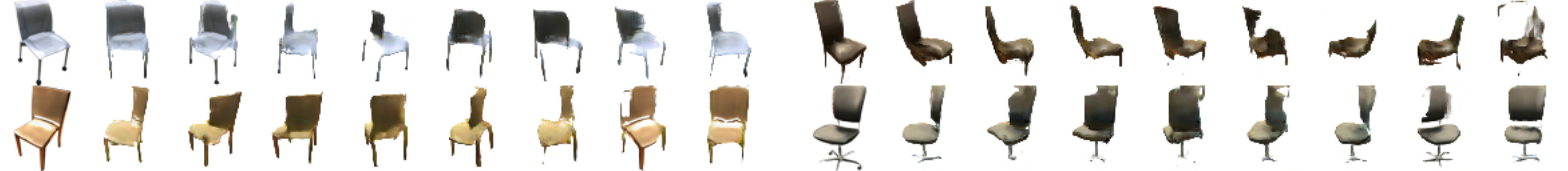} \\
			\hline
			\textbf{DRAW} & \includegraphics[width=0.85\textwidth]{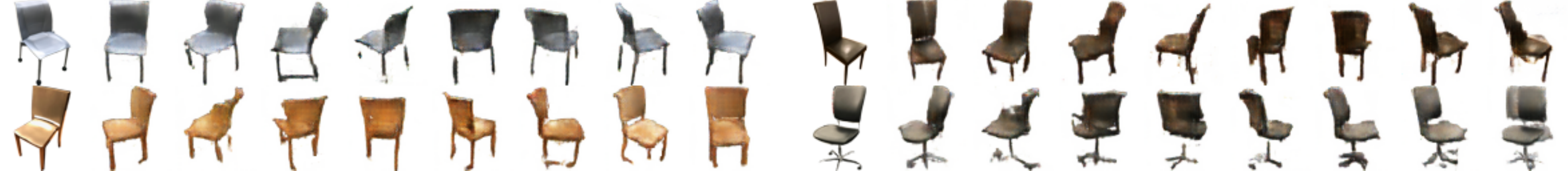}
		\end{tabular}}
	\caption{View synthesis comparisons on \emph{chair} images from Pix3D. Only 9 out of 18 views are shown. Note that~\cite{sun2018multi} is trained and tested with multiple views. DRAW generates the entire trajectory with a single image (best-viewed zoomed).}
	\label{fig:overall_figures}
\end{figure*}

\begin{figure*}[t!]
\setlength{\belowcaptionskip}{-0.5cm}
\centering{
		\begin{tabular}{c|ll}
			\cite{tatarchenko2016multi} & \includegraphics[width=0.40\textwidth]{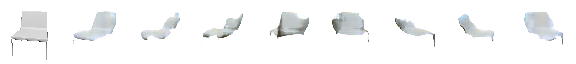} & \includegraphics[width=0.40\textwidth]{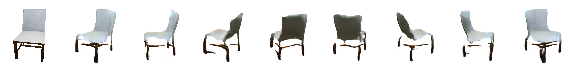} \\
			\hline
			\cite{zhou2016view} & \includegraphics[width=0.40\textwidth]{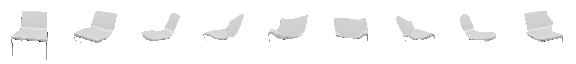}
			 & \includegraphics[width=0.40\textwidth]{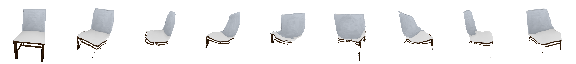} \\
			\hline
			\cite{sun2018multi} & \includegraphics[width=0.40\textwidth]{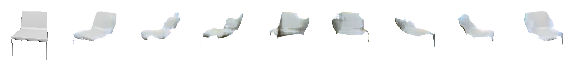}
			& \includegraphics[width=0.40\textwidth]{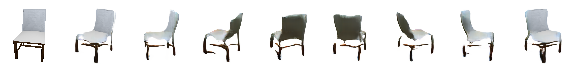} \\
			\hline
			\textbf{DRAW} & \includegraphics[width=0.40\textwidth]{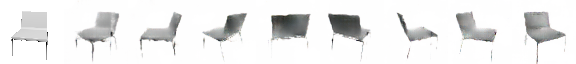} &
			\includegraphics[width=0.40\textwidth]{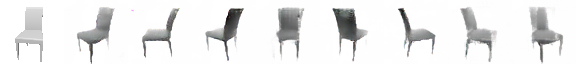}
		\end{tabular}}
	\caption{Comparison of view synthesis results on \emph{chairs} from ShapeNet (best-viewed zoomed).}
	\label{fig:shapenet_figures}
\end{figure*}

\begin{figure}[t!]
\setlength{\belowcaptionskip}{-0.5cm}
	\centering
	\includegraphics[width=0.43\textwidth]{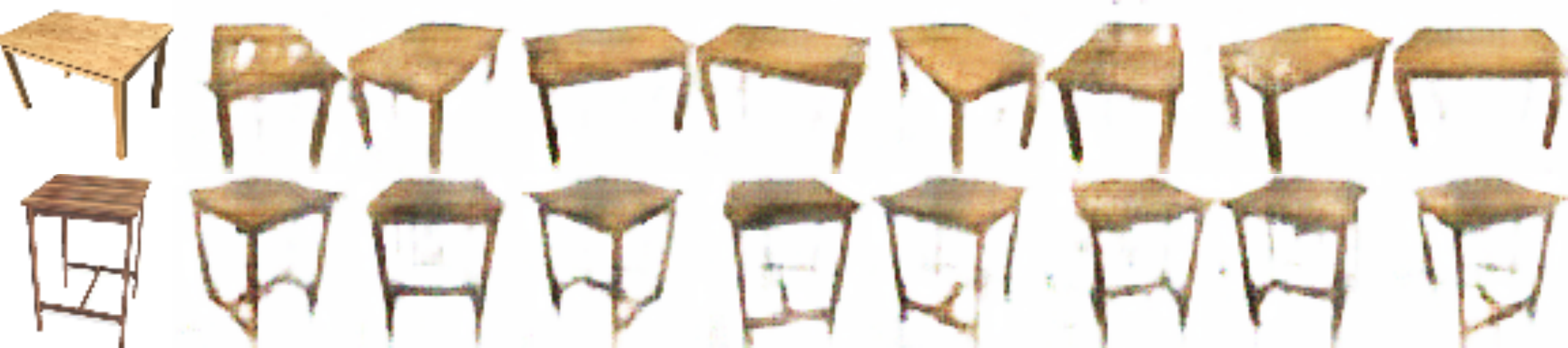}
	\caption{DRAW view synthesis on \emph{table} images from Pix3D.}
	\label{fig:table}
\end{figure}

\vskip0.5ex
\noindent
{\bf Identity recovery (IR).} We consider two baselines and a loss function variation for comparison.
The model in Fig.~\ref{fig:recovery_variation} (\emph{top})
simply treats the problem as one of image to image translation. Since
it only has access to the depth map $\mathbf{s}_p$, it has to
hallucinate the object appearance.
We refer to it as the {\it hallucination\/}
(\texttt{HAL}) model. The model of Fig.~\ref{fig:recovery_variation} (\emph{bottom}) 
is a simpler variant of the
identity recovery module of Fig.~\ref{fig:identity}. It has access
to both $\mathbf{x}_0$ and $\mathbf{s}_p$, but imposes much weaker
disentanglement constraints because it does not require
the synthesis of all combinations of shape and appearance. We refer to
it as the {\it weak identity recovery\/} (\texttt{WIR}) module. 
A loss function variation is applied to our proposed model. For the training of our identity recovery model, we now include $\mathbf{x}_p$ in training and alter the supervised loss of Eq.~\eqref{eqn:irdisc} to
$
{\cal L}^{\texttt{S}}_{\texttt{IR}} = \mathbb{E}_{\mathbf{x}_0,\mathbf{s}_0,\mathbf{x}_p,\mathbf{s}_p}[\|\mathbf{x}_0-\mathbf{\hat{x}}_0\|_1 + \|\mathbf{s}_0-\mathbf{\hat{s}}_0\|_1
 + \|\mathbf{x}_p-\mathbf{\hat{x}}_p\|_1 + \|\mathbf{s}_p-\mathbf{\hat{s}}_p\|_1]
$. This loss is not possible during training the full model end-to-end, but can be done when training the \texttt{IR} model separately. Our goal is to examine how much information we lose by removing the direct supervision on $\mathbf{x}_p$. From Fig.~\ref{fig:recovery_variation}, we see that $L_1$ and SSIM do not change much by removing $\mathbf{x}_p$ from supervision.

All models are trained on pairs of RGB-D images corresponding to different
views of the same object in Pix3D. During inference, a RGB image from
view $1$ and a depth map from view $2$ are used to predict a RGB image from
view $2$. Due to the lack of supervision for target RGB
images, \texttt{HAL} and \texttt{WIR} are optimized using the adversarial loss alone.
Unsurprisingly, \texttt{HAL} has weak performance. 
It is also clear that the additional
disentanglement constraints of \texttt{IR} lead to a performance improvement over
\texttt{WIR}.
Fig.~\ref{fig:recovery_results} (\emph{top}) shows  examples of the views
synthesized by the identity recovery module of DRAW. Note the quality of synthesis across
very large view angle transformations.  Fig.~\ref{fig:recovery_results}
(\emph{bottom}) further shows recovery results where two objects of identical shape but different appearance are presented. This example clearly 
demonstrates that the identity recovery module produces outputs that carry the 
input images' appearance upon receiving a common $\mathbf{s}_p$ but different 
$\mathbf{x}_p$.

\subsection{Comparison to the state-of-the-art}

We now evaluate the entire DRAW pipeline
on the original task of {\it pose trajectory learning from sparse image data\/}.
For this experiment we use the Pix3D dataset as well a subsampled version of the ShapeNet dataset, 
where, for each object instance, only $\sim$5 of its poses are retained in RGB, for training. We compare DRAW with
the state-of-the-art novel view synthesis methods~\cite{tatarchenko2016multi,zhou2016view,sun2018multi}
in this sparse setting. For evaluation on Pix3D, the pre-trained models from~\cite{tatarchenko2016multi,zhou2016view,sun2018multi}
are fine-tuned to the Pix3D training data. For the ShapeNet experiment, these models are trained on the sub-sampled dataset
from scratch.

\vskip0.5ex
\noindent
\textbf{Qualitative analysis on Pix3D.}
Qualitative results of all methods are shown in Fig.~\ref{fig:overall_figures}. DRAW results for \emph{tables} can be found in Fig.~\ref{fig:table}.
\begin{table}[t!]
\setlength{\belowcaptionskip}{-0.5cm}
	\begin{footnotesize}
		\centering
		\begin{tabular}{rcccc}
			\toprule
			& \cite{tatarchenko2016multi} & \cite{zhou2016view} & \cite{sun2018multi} & \textbf{DRAW} \\
			\hline
			\multicolumn{5}{c}{Pix3D} \\
			\midrule
			$L_1$					 & 0.16 & 0.15 & 0.16 & \cellcolor{green!20}{{\bf 0.12}} \\
			SSIM					 & 0.45 & 0.46 & 0.45 & \cellcolor{green!20}{{\bf 0.51}} \\
			\hline
			\hline
			\multicolumn{5}{c}{ShapeNet} \\
			\midrule
			$L_1$					 & 0.24 & 0.25 & 0.24 & \cellcolor{green!20}{{\bf 0.20}} \\
			SSIM					 & 0.49 & 0.46 & 0.47 & \cellcolor{green!20}{{\bf 0.56}} \\
 			\bottomrule
		\end{tabular}
		\caption{Cross-domain view synthesis comparison on Pix3D. For 
		$L_1$ \textbf{lower} is better, for SSIM and inception, \textbf{higher}
		is better.
		\label{tab:overll_results}}
	\end{footnotesize}
\end{table}
DRAW appears to preserve the appearance and the structure of the object across different poses. Other methods not do not seem to perform very well in comparison. This can be attributed to the severe deficiency of pose training data in Pix3D. Fine-tuning the models of~\cite{tatarchenko2016multi,zhou2016view,sun2018multi} to limited object views leads to poor generalization. Notably, the results of~\cite{zhou2016view} are the worst among all methods. This is perhaps due to the lighting variations and realistic texture in natural images, which results in poor appearance flow mapping compared to synthetic domain. It is also noteworthy that previous methods require view point labels for training. \emph{DRAW doesn't need such supervision}.
   
\vskip0.5ex
\noindent
\textbf{Quantitative analysis on Pix3D.} We compiled a test set of valid trajectories from Pix3D, to evaluate the
reconstructions of DRAW and it's competitors~\cite{tatarchenko2016multi,zhou2016view,sun2018multi} trained with sparse pose supervision.
When one image in the trajectory is used as input, all other images are used as targets to calculate the errors between the related generated view. Results for $L_1$ and SSIM measures, listed in Table~\ref{tab:overll_results}, indicate that DRAW outperforms all other methods.

\vskip0.5ex
\noindent
\textbf{Analysis on sparse ShapeNet.}
We next compare all methods on a pose-sparse subset of Shapenet constructed by randomly sampling $5$ RGB views per object instance. Models for~\cite{tatarchenko2016multi,zhou2016view,sun2018multi} are trained from scratch on this subsampled data. Qualitive images are shown in Fig.~\ref{fig:shapenet_figures} and quantitative results in Table~\ref{tab:overll_results} (bottom). It seems clear that in case of sparse pose training, DRAW performs better than other methods even on synthetic ShapeNet. 

\section{Conclusion}
We introduce DRAW, a framework to learn pose trajectories from natural image datasets with very few canonical views available per object instance. To make up for such sparsity in the image space, we propose the use cross-modal guidance in the form of texture-less 3D CAD models available for objects. Dense pose trajectories of depth maps can be extracted from these models. Given an RGB image, DRAW operates by (1) mapping it into 2D depth maps, (2) simulating 3D object rotation in depth space and (3) re-mapping the generated views from depth to image space. DRAW can be trained with a set of images with \emph{sparse views}, as in Pix3D. Pose trajectories are synthesized in the synthetic domain and transferred to the image domain in a manner that guarantees consistency of object identity. An identity recovery network that disentangles and recombines appearance and shape information was shown to be important for this purpose. Experiments show that state-of-the-art pose synthesis methods do not perform well if trained with sparse object views in image space. DRAW overcomes this issue by leveraging cross-modal structure priors and generates results of reasonable quality, structural integrity and instance identity.


{\small
\bibliographystyle{ieee_fullname}
\bibliography{view}
}

\end{document}